\documentclass[11pt]{article}
\usepackage{enumitem}
\usepackage{float}
\usepackage{amssymb}
\usepackage{makecell}
\usepackage{bm}
\usepackage{graphicx}
\usepackage{listings}
\usepackage{xcolor}
\usepackage{adjustbox}
\usepackage{booktabs}
\usepackage{multirow}
\usepackage{graphicx}
\usepackage{caption}
\usepackage{subcaption}
\usepackage{float}

\usepackage[final]{acl}
\usepackage{amsmath}

\usepackage{times}
\usepackage{latexsym}

\usepackage[T1]{fontenc}

\usepackage[utf8]{inputenc}
\usepackage{newunicodechar}
\newunicodechar{’}{'}
\newunicodechar{–}{--}
\newunicodechar{—}{---}
\newunicodechar{“}{``}
\newunicodechar{”}{''}

\usepackage{microtype}

\usepackage{inconsolata}

\usepackage{graphicx}

\title{Labels have Human Values: Value Calibration of Subjective Tasks}

\author{Mohammed Fayiz Parappan \quad Ricardo Henao \\
        Department of Electrical and Computer Engineering \\
        Duke University \\
        \texttt{\{mohammedfayiz.parappan, ricardo.henao\}@duke.edu}}

\begin{document}
\maketitle
\begin{abstract}
Although pluralistic societies exhibit diverse human values that lead to legitimate disagreements in subjective tasks ({\em e.g.}, safety and preference judgments), NLP models trained on such subjective labels often ignore latent value structures, resulting in miscalibrated predictions over relevant value classes.
We propose MultiCalibrated Subjective Task Learning (MC-STL)\footnote{
Implementation available at \url{https://github.com/mohammedfayizp/MC-STL_code}.
}, a framework that identifies latent value groups from annotations (via label rationale similarity, expert value taxonomies, or annotator sociocultural descriptors) and enforces value-conditional calibration through value group-specific representations.
MC-STL applies to binary, ordinal, and preference learning settings, and is evaluated on multiple datasets covering toxic chatbot conversations, value reasoning, T2I-safety and preference alignment.
The results demonstrate that MC-STL consistently outperforms existing baselines, achieving multicalibration across relevant value groups, while delivering gains in probabilistic prediction performance.

\end{abstract}


\section{Introduction}
\begin{figure}[H]
  \centering
  \includegraphics[width=\linewidth]{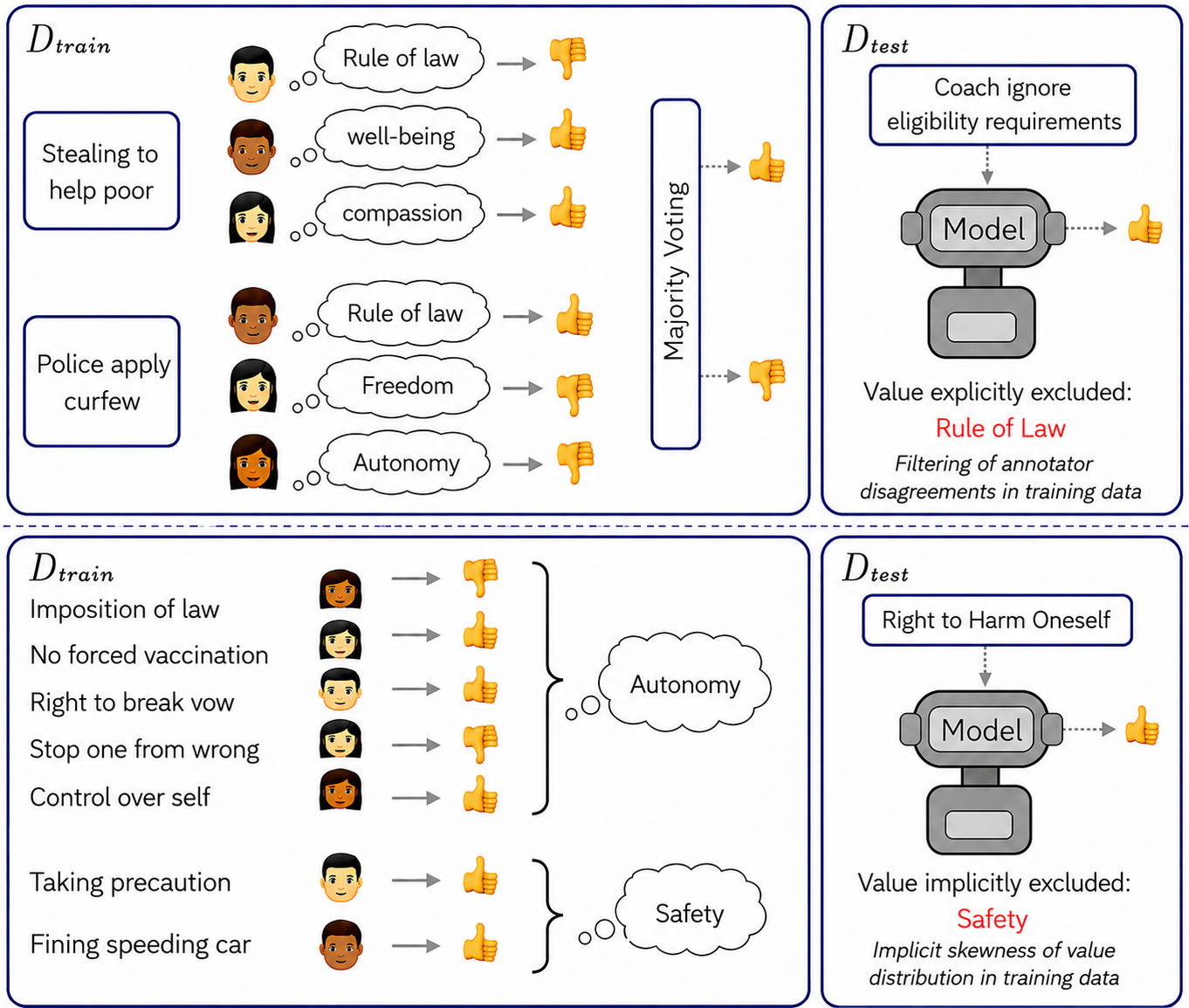}
  \caption{Example of how subjective label predictions are affected by hidden human values (denoted as thought clouds). (Top) when annotator disagreements are filtered out by majority voting or (Bottom) when minority values are underrepresented and likely unaccounted by the model in crowd-sourced annotation settings.
  }
  \label{fig:fullwidth}
  \vspace{-3mm}
\end{figure}

As NLP systems are increasingly deployed in socially sensitive settings, ensuring reliable behavior across the diversity of human values has become an important challenge \citep{pehlivan2024eu, un2023global, sorensen2024roadmap}.
However, conventional machine learning workflows often suppress value diversity, either explicitly through majority aggregation \citep{hovy2013learning} or implicitly when learning methods average over latent contexts in training data that are skewed towards certain perspectives unknown to the model but overrepresented in the data \citep{obi2024value, siththaranjan2023distributional}; as illustrated in Figure~\ref{fig:fullwidth}.

Ignoring the latent value structure underlying subjective annotations can lead to systematic misalignment in high-stakes decision systems \citep{christian2026reward, elle-2025-reward}, and recent work has called for explicit modeling of subjective value perspectives behind annotations \citet{kirk2024prism}.

Following \citet{schwartz2012overview}, subjectivity in human judgements often arises from differing underlying {\em human values}. We build on the intersection of subjective NLP and value alignment by identifying distinctive {\em human value groups} from annotations and learning reliable predictions across these groups. The framework, hence, could also disambiguate annotation disagreements arising from value conflicts.
In terms of {\em value alignment}, we specifically focus on multicalibration \citep{hebert2018multicalibration, corvelo2023human} of a subjective NLP system to relevant human value groups.
Value group multicalibration requires prediction probabilities to remain calibrated across relevant value group perspectives rather than only globally calibrated, thereby ensuring reliable probabilistic predictions for diverse human viewpoints.
To the best of our knowledge, the proposed MultiCalibrated Subjective Task Learning (MC-STL) is the first framework to explicitly model latent human value group perspectives underlying subjective annotations while enforcing calibration over all value groups.
Our contributions are as follows.

\begin{itemize}[itemsep=-1mm,topsep=-1mm,leftmargin=3mm]
    \item The MC-STL framework identifies latent value groups among subjective annotations by label rationale clustering, expert-defined value taxonomy mapping, or sociocultural descriptors, and learns calibrated predictions over these value groups.
    \item The MC-STL framework supports various task scenarios, such as binary classification, ordinal classification, and human preference learning. Moreover, it is scalable to diverse, multimodal subjective applications such as toxicity detection, value reasoning, T2I (Text-to-Image) Safety, {\em etc}.
    \item Across multiple subjective NLP benchmarks, MC-STL improves predictive performance, achieves more reliable multi-calibration over value groups, and better preserves minority viewpoints compared to existing subjective modeling baselines and value agnostic methods.
\end{itemize}

\section{Related Work}
\label{sec:related}
\textbf{Subjectivity in NLP: }Computational linguists have long emphasized the importance of modeling subjective meaning and the individual or social dimensions of language \citep{wiebe2004learning, wilson2005recognizing}. These dimensions are central to many modern NLP tasks that are inherently subjective, such as hate speech detection \citep{akhtar2020modeling, warner2012detecting}, sentiment analysis \citep{liu2010sentiment, kenyon2018sentiment} and human preference modeling \citep{elle-2025-reward}. However, traditional data curation pipelines for such tasks often treat annotator disagreement as noise, resolving it through majority voting or averaging \citep{hovy2013learning, sabou2014corpus}. This challenge extends beyond observable disagreement: most NLP models are trained on crowdsourced data where latent contexts—such as annotator values, identities, or interpretive frames, which guide labeling behavior but are not visible to the model—vary widely. The accumulation of a dominant hidden context can bias models away from pluralistic alignment \citep{obi2024value, siththaranjan2023distributional, sorensen2024roadmap}. \citet{atari2023humans} found AI values to be WEIRD (Western, Educated, Industrialized, Rich and Democratic) and disregarding pluralistic values on subjective topics.

\textbf{Modeling Subjectivity: }
Annotator disagreement is often treated as an informative signal rather than annotation noise to model subjectivity
\cite{prabhakaran2021releasing}.
One line of work attempts to model individual annotators directly, using learnable annotator-specific embeddings or classification heads \citep{gordon2021disagreement, davani2022dealing, hayat2022modeling, mokhberian2024capturing}. Although effective in capturing annotator-specific variation, these methods struggle to generalize to unseen users. \citet{vitsakis2024voices} identify unique perspective clusters among annotators by clustering annotator specific learned representations. The approach assumes clusters of annotators sharing similar perspectives in all cases. Another approach models subjectivity using annotator descriptors—such as age, gender, ethnicity, or political affiliation—known to correlate with labeling behavior \citep{talat2016you,luo2020detecting,prabhakaran2021releasing,fleisig2023majority,gordon2022jury}. More recently, \citet{parappan2025learning} leverage sociocultural descriptors to predict calibrated distributions of binary toxicity judgments on a text item. The method requires substantial annotation per text item from demographically diverse annotators, making it costly in practice. 

\textbf{Value Modeling in Subjective Task: } Current NLP alignment approaches focus primarily on post-training alignment of generated outputs through activation tuning, reward modeling, or value-space mapping \citep{jin2025internal, christiano2017deep, yao2024value}. However, these methods primarily address the alignment of content generated by LLMs and are usually not extendable to NLP systems designed for categorical decision tasks. 

According to \citet{schwartz2012overview}, human values encode desirable goals guiding behavior (here, annotation).
Recent datasets, {\em e.g.}, \citet{kirk2024prism, sorensen2024value} include rater rationales linked to annotations. \citet{obi2024value} introduce a framework for identifying human value classes embedded in RLHF datasets. \citet{sorensen2025value} demonstrate that the value profiles derived from summarizing annotators’ justifications (by LLMs) provide effective representations of individual rater perspectives. However, such annotator-centric modeling suffers from scaling to unseen populations.
 
Building on these developments, our work jointly models latent value groups underlying subjective annotations and learns calibrated predictions over these value groups.

\begin{figure*}[t]
  \centering
  \includegraphics[width=\textwidth, height=4.5cm, keepaspectratio=false]{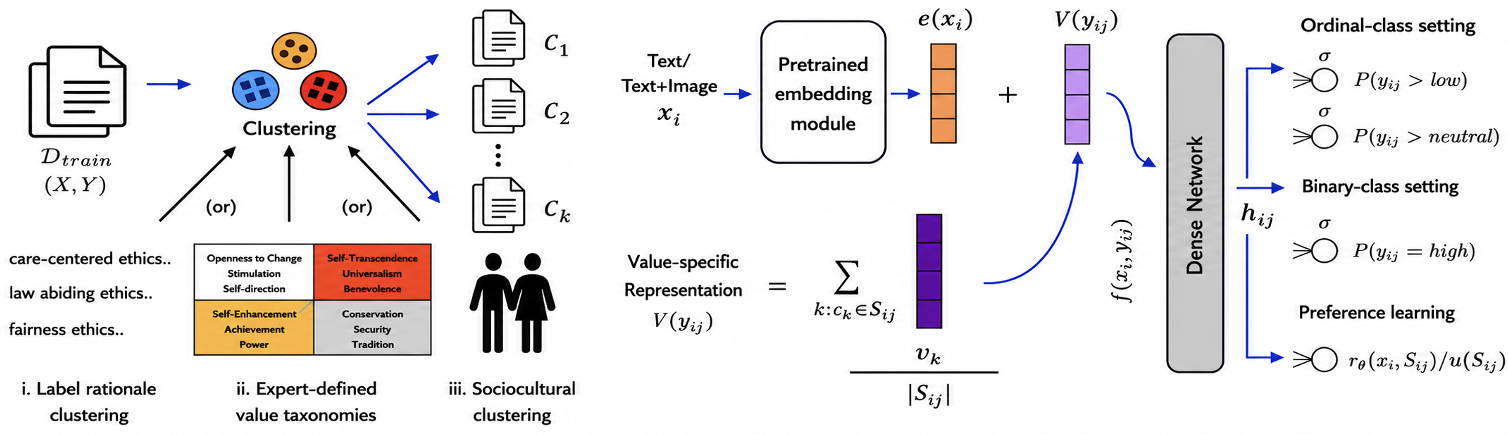}

  \hspace*{-0.12\textwidth}(a)\hspace{0.52\textwidth}(b)

  \caption{Overview of the proposed MC-STL framework.
    (a) Identification of value groups.
    (b) MC-STL model architecture.}
  \label{fig:overview}
  \vspace{-4mm}
\end{figure*}


\section{Methodology}
\subsection{Problem Definition}
Let $\mathcal{D} = (\mathbf{X}, \mathbf{Y})$ denote an annotated dataset for a subjective natural language processing (NLP) task.
The dataset consists of a collection of $N$ text items $\mathbf{X} = \{x_i\}_{i=1}^N$, where each item $x_i$ is associated with a set of annotations $\mathbf{y}_i = \{y_{ij}\}_{j=1}^{m_i}$, with $m_i$ denoting the number of annotations collected for the item $x_i$.
The collection of annotation sets $\mathbf{Y} = \{\mathbf{y}_i\}_{i=1}^N$ constitutes the set of label annotations for the complete dataset.
The annotations $y_{ij}$ can take values from binary or ordinal label spaces.
The task is twofold: first, to map each annotation $y_{ij}$ to one or more of its relevant {\em human value groups} (by clustering with commonly available meta-data or expert analysis).
Let $\mathcal{C} = \{ c_k \}_{k = 1}^K$ denote the resulting set of $K$ value groups obtained by clustering annotations, where each value group $c_k$ is a subset of annotations sharing a common value-driven perspective. 
Value groups may follow a mixed membership model, where an annotation $y_{ij} \in {\bf Y}$ can belong to more than one value group in $\mathcal{C}$.
Let $S_{ij}$ be the set of value groups to which $y_{ij}$ is assigned, then $S_{ij}=\{c_k \in C \mid y_{ij} \text{ is assigned to } c_k\}$.
The second task is to learn a model whose predictions $\hat{P}\!\left( y_{ij} \mid x_i,\ S_{ij} \right)$ are multi-calibrated over value groups, $c_k \in C$.

\subsection{MC-STL Framework}
\subsubsection{Value Group Construction}

We consider three approaches for the construction of value groups in $C$ from subjective annotations, capturing distinct notions of value group structures, as described in the following.

\noindent\textbf{Label rationale clustering}
Given free-text rationales or value statements associated with each annotation $y_{ij}$, either collected during annotation or retrospectively inferred using LLMs \citep{sorensen2025value, hayati2024far}, we cluster the annotations in the train split of $D$ using $K$-means over sentence embeddings of their rationales.
The resulting clusters group annotations that share semantically similar value themes.
Each value group (cluster) may then be interpreted by summarizing its associated rationales using either an LLMs or domain experts.
The number of value groups $K$ can be selected through interpretability analysis by inspecting value-group summaries or by validation performance. 
At test time, the embedding of the rater rationale for an instance is mapped to its closest cluster centroid, and the instance is evaluated based on the corresponding value group.


%
%

\noindent\textbf{Expert-defined value taxonomies} 
We also consider predefined human value taxonomies that directly map each annotation $y_{ij}$ to established value classes (either during annotation itself or later by mapping associated free-text rater rationales to value groups by expert), forming value groups among their annotations.
For example, \citet{schwartz2012overview} defines ten universal human value classes.
Similar taxonomies are widely used in RLHF settings \cite{bai2022training}, where preference annotations reflect high-level value dimensions such as \textit{Helpfulness} and \textit{Harmlessness}.
In this setting, $K$ corresponds to the number of predefined value classes.
%


\noindent\textbf{Sociocultural Clustering} 
In some subjective tasks, annotation differences are influenced by sociocultural factors \citep{aroyo2023dices, dev2026unified, davani2024disentangling} and ensuring fairness over all sociocultural groups is a critical concern.
Here, annotations are grouped according to annotator attributes such as age, ethnicity, or geographic location.
Although sociodemographic attributes of the annotator do not directly represent value groups, they can be used to form groups in tasks where differences in subjective judgments are associated with sociocultural factors. In this setting, $K$ equals the number of sociocultural groups considered and the groups act as proxies for latent cultural value perspectives rather than direct representations of human values. 


The three approaches above differ from each other in terms of the value groups formed.
Rationale clustering infers latent value groups directly from label rationale semantics and is particularly useful when task-relevant value distinctions are unknown {\em a priori}.
In contrast, expert-defined taxonomies provide task-independent value dimensions aligned with established human value theories, while sociocultural clustering captures demographic and cultural variation underlying subjective judgments.
Choosing a method for the construction of value groups for a task therefore depends on the task objectives, fairness considerations, and data availability.

\subsubsection{MC-STL Model Architecture}
The proposed \text{MC-STL} model adopts a value-aware architecture to model human subjectivity using the constructed value groups, as illustrated in Figure~\ref{fig:overview}.
Given a text item $x_i$, we first encode it using a pretrained transformer-based encoder into a $d$-dimensional embedding $e(x_i) \in \mathbb{R}^d$.

For each value group $c_k \in \mathcal{C}$, we initialize a unique learnable embedding $v_k \in \mathbb{R}^d$.
These embeddings are trained jointly with the rest of the model parameters and are intended to capture value-specific inductive biases.
For a given annotation $y_{ij}$, the corresponding value-specific representation is obtained by averaging the embeddings of all value groups associated with it as follows.
\begin{align}\label{eq:v}
    V(y_{ij}) = \frac{\textstyle{\sum}_{k \,:\, c_k \in S_{ij} } v_k }{|S_{ij}|}\,.
\end{align}
For each training instance $(x_i, y_{ij})$, we construct a value-aware representation by combining the text embedding with the value-specific embedding as
$
f(x_i, y_{ij}) = e(x_i) + V(y_{ij}) 
$.
%
This formulation naturally supports value group mixed membership (such as in Schwartz's value theory) while preserving the dimensionality of $f(x_i, y_{ij})$ through \eqref{eq:v}.

The value-aware representation $f(x_i,y_{ij})$ is passed to a feed-forward prediction network. Let
$h_{ij}$
denote the scalar output of the dense transformation excluding final-layer bias parameters.
We employ task-specific output heads depending on the prediction setting:
$i$) ordinal label setting to accommodate subjective annotations with ordered response scales ({\em e.g.}, Likert-style labels) \citep{sorensen2024value, aroyo2023dices}, and
$ii$) a binary label setting for explicit two-class scenarios.

\paragraph{Ordinal classification setting}
For ordinal prediction tasks, {\em i.e.}, with \(r\) ordered categories, we adopt the CORAL (Consistent Rank Logits) framework \citep{cao2020rank}, which decomposes ordinal prediction into \(r-1\) binary thresholding sub-tasks.

Let \(y_{ij} \in \{l_1, l_2, \dots, l_r\}\) denote an ordinal label space with ordered categories \(l_1 < l_2 < \dots < l_r\). For each threshold \(t \in \{1,\dots,r-1\}\), CORAL defines a binary target $y_{ij}^{(t)} = \mathbf{1}[y_{ij} > l_t]$.
The probabilities for each target are defined as
\[
\hat{p}_{ij}^{(t)}
=
\hat{P}(y_{ij}^{(t)} = 1)
=
\sigma\left(h_{ij} + b_t\right)
\]
where \(t \in \{1,\dots,r-1\}\) indexes the ordinal thresholds, $b_t$ is the bias parameter associated with threshold $t$ and $\sigma(\cdot)$ is the sigmoid function.
Following CORAL, this formulation enforces rank consistency across ordinal thresholds.
The class probabilities are recovered from the adjacent cumulative probabilities following the standard CORAL formulation \citep{cao2020rank}.

\paragraph{Binary classification setting}
It corresponds to the special case where $r=2$. Given labels \(y_{ij} \in \{0,1\}\), the model reduces to a single prediction head $\hat{p}_{ij}=\hat{P}(y_{ij}=1)=\sigma\left(h_{ij} + b\right)$.
\subsubsection{Loss Function}
Our training objective is designed to jointly optimize:
$i$) probabilistic prediction at the instance level, and
$ii$) distributional calibration at the value group level.
For ordinal prediction tasks with \(r\) ordered categories, the model produces \(r-1\) binary target predictions following the CORAL formulation described above.
For each threshold \(t \in \{1,\dots,r-1\}\), we optimize a binary cross-entropy (CE) objective and a group-level distribution matching regularizer based on the Kullback-Leibler (KL) divergence.
The complete training objective is
%
\begin{align}
\mathcal{L}_{\mathrm{ord}}
=&\sum_{i,j,t}\mathcal{L}_{\mathrm{CE}}
\!\left(y_{ij}^{(t)},\hat{p}_{ij}^{(t)}\right) \label{loss_ordinal} \\
+&\lambda_1\sum_{k,t}\alpha_k
\mathrm{KL}\!\left(P_k^{(t)}\|Q_k^{(t)}\right) 
+\lambda_2\sum_k\|v_k\|_2^2 , \notag
\label{loss_ordina}
\end{align}
where
\(
\mathcal{L}_{\mathrm{CE}}
\)
denotes binary CE between the target
\(
y_{ij}^{(t)}
\)
and prediction
\(
\hat{p}_{ij}^{(t)}
\), the KL divergence encourages the average predicted distribution within each value group \(c_k \in \mathcal{C}\) to match the corresponding empirical label distribution from the training data, and the last term regularizes the embeddings, $\{v_k\}_{k}$.

Let
\(
P_k^{(t)}
\)
and
\(
Q_k^{(t)}
\)
denote the empirical and predicted Bernoulli distributions, respectively, for each value group \(c_k\) and threshold \(t\).
We define
%
\begin{equation*}
\begin{aligned}
\mathrm{KL}(P_k^{(t)} \| Q_k^{(t)})
= & \ n_k \bar{y}_k^{(t)}
\ln\!\left(
\frac{\bar{y}_k^{(t)}}{\hat{p}_{k}^{(t)\prime}}
\right) \\
+ & \ n_k\!\left(1-\bar{y}_k^{(t)}\right)
\ln\!\left(
\frac{1-\bar{y}_k^{(t)}}{1-\hat{p}_{k}^{(t)\prime}}
\right).
\end{aligned}
\end{equation*}
%
where \(n_k\) is the number of annotations associated with value group \(c_k\),
\(
\bar{y}_k^{(t)}
\)
is the empirical mean of
\(
\{y_{ij}^{(t)}\}_{ij \in c_k}
\),
and
\(
\hat{p}_{k}^{(t)\prime}
\)
is the corresponding average predicted probability.

Although we have discrete realizations $y_{ij}\in\{0,1\}$ from $P_k$ to obtain $\bar{y}_k$, we only have predicted probabilities for the realizations of $Q_k$.
We soft-threshold predictions wrt the empirical mean $\bar{y}_k^{(t)}$ before aggregating them to obtain $\hat{p}_{k}^{(t)\prime}$.
Specifically, we calculate $\hat{p}_{k}^{(t)\prime}$ after converting each probability into {\em approximately} binary labels using
%
\begin{equation*}
\hat{p}_{k}^{(t)\prime}
=
\frac{1}{2}
\frac{
\sum_{{ij} \in C_k}
\left(
1 + \tanh\!\left(
l \cdot (\hat{p}_{ij}^{(t)} - \bar{y}_k^{(t)})
\right)
\right)
}{n_k}
\,.
\end{equation*}
%
where the hyperbolic tangent (tanh) function, with a large constant $l=10^4$ serves as a relaxation to the use of predictions with hard thresholds while allowing for smooth gradient flow during training.

In \eqref{loss_ordinal}, $\|v_k\|$ is an ($L_2$) regularizer for the embedding associated with the value group $c_k \in {\cal C}$.
Note that $v_k$ is high-dimensional and matches the dimensionality of $e(x_i)$, which is fixed.
Moreover, $\alpha_k$ is a parameter that weights $\mathrm{KL}(P_k^{(t)}\|Q_k^{(t)})$ according to whether their label distribution $P_k^{(t)}$ matches that of the training set, thus up-weighting value groups for which the distributions are different.
Details are presented in Appendix~\ref{app:KL}.
Finally, $\lambda_1$ and $\lambda_2$ are hyperparameters that balance the contributions of KL divergences and L2 regularizer, which are determined by grid search.

\noindent{\bf Binary setting}
It is a special case of \eqref{loss_ordinal} with \(n=2\) requiring a single threshold, thus reducing it to
%
\begin{equation*}
\label{loss_binary}
\begin{aligned}
\mathcal{L}_{\mathrm{bin}}
= & \sum_{i,j}
\mathcal{L}_{\mathrm{CE}}(y_{ij}, \hat{p}_{ij}) \\
+ & \lambda_1 \sum_{k} \alpha_k
\mathrm{KL}(P_k \| Q_k)
+ \lambda_2 \sum_{k}\| v_k \|_2^2 .
\end{aligned}
\end{equation*}
\subsection{Extension to Preference Learning}
Preference learning settings commonly used in RLHF-based reward modeling \citep{christiano2017deep} require pairwise preference annotations that often reflect latent human values.
Preference confidence scores are frequently used to filter synthetic labels \citep{Shen2024PolicyFF} without explicitly modeling calibration over distinct value groups.

We extend MC-STL to preference learning as follows.
%
%
Let $\mathcal{D}=\{(x_i,x_j,y_{ij})\}$
denote a dataset of pairwise preferences, where $y_{ij}\in\{0,1\}$ indicates whether text item $x_i$ is preferred over $x_j$.
Each preference annotation is associated with a set of latent value perspectives $S_{ij}\subseteq C$. Given a text item $x_i$ and value assignment $S_{ij}$, the MC-STL backbone as in Figure~\ref{fig:overview} produces a scalar reward score
\(
r_\theta(x_i,S_{ij})=h_{ij}
\).
To model value-dependent uncertainty in preference judgments, we introduce a learnable temperature parameter $u_k$ for each value group.
For mixed value group assignments, we define the effective temperature as
\[
u(S_{ij})
=
\frac{1}{|S_{ij}|}
\sum_{c_k\in S_{ij}} u_k.
\]
The probability that $x_i$ is preferred over $x_j$ is then modeled using a Bradley--Terry formulation:
\[
\hat{p}(y_{ij})
=
\sigma\left(
\frac{
r_\theta(x_i,S_{ij})
-
r_\theta(x_j,S_{ij})
}{
u(S_{ij})
}
\right),
\]
and the corresponding training objective is
\[
\mathcal{L}_{pl}
=
\sum_{i,j}
\mathcal{L}_{BT}(y_{ij},\hat{p}(y_{ij}))
+
\lambda_2
\sum_k
\|v_k\|_2^2,
\]
where $\mathcal{L}_{BT}$ denotes the Bradley--Terry preference loss \cite{christiano2017deep}.

\section{Experiments}
All experiments were conducted on a server equipped with a single NVIDIA RTX A6000 GPU.
For label rationale clustering, we used the
GPT-4o API to summarize rationales in each value group (prompt details are provided in Appendix~\ref{sec:prompt_sum}).
For text-only items (excluding preference learning), text representations $e(x_i)$ were obtained using pretrained Sentence-BERT embeddings (all-MiniLM-L6-v2) \citet{reimers2019sentence}.
For preference learning tasks, we used Longformer-base-4096 \citep{beltagy2020longformer} to accommodate long multi-turn context input. 
The model was finetuned for 2 epochs on the training set, and the final hidden-state representation of the [CLS] token was used as the text embedding.
For text-to-image (T2I) safety, joint prompt-image representations were constructed by summing the pretrained CLIP embeddings \cite{radford2021learning} of the text prompt and the generated image.
The evaluation was performed on a (held-out) test split consisting of 20\% of each dataset.
Data splits were constructed using stratified sampling to preserve label distributions.
Additional details are provided in Appendix~\ref{sec:add_exp}.

\subsection{Datasets}
We evaluate MC-STL over subjective NLP tasks spanning value reasoning, toxicity detection, RLHF preference modeling, and multimodal safety.

\noindent{\bf ValuePrism}
\citep{sorensen2024value} is a large-scale subjective reasoning dataset that contains approximately 31k text situations paired with 218k annotations.
Each annotation includes an ordinal label (\textit{Oppose}, \textit{Either}, or \textit{Support}) together with a free-text rationale. We evaluate MC-STL separately on two rationale categories: {VP-Rights} and {VP-Duty}, enabling experiments under distinct normative framings.

\noindent\textbf{ValuePrism + Schwartz (VP-Schwartz) Value Labels}
To evaluate expert-defined value taxonomies, we extend ValuePrism with $10$ Universal Schwartz value class labels \cite{schwartz2012overview}.
We map annotations to one or more Schwartz value classes using GPT-4o-mini following the calibrated Schwartz value attributor prompt proposed by \cite{guo2026counterfactual}, and a random subset (3k samples) of each $c_k$ formed was further verified by human experts (psychologists), achieving 95\% agreement with the generated mapping.
See Appendix~\ref{sec:expert} for additional details.

\noindent\textbf{Anthropic HH-RLHF}
\cite{bai2022training} is a widely used preference-learning dataset that contains approximately 169k human preference pairs.
We use the ValueImprint annotations \citep{obi2024value}, which associate preference pairs with high-level human value classes.      

\noindent\textbf{DICES-990}
\cite{aroyo2023dices} is a dataset of 990 conversations, each annotated for toxicity by 60 raters using ordinal labels: \textit{Unsafe}, \textit{Unsure}, and \textit{Safe}.
Each annotation is linked to annotator demographics, {\em i.e.}, gender, race, age group, and locality.

\noindent\textbf{DIVE}
\citep{rastogi2026whose} is a multimodal text-to-image (T2I) safety dataset containing 1000 prompt-image pairs annotated for safety by demographically diverse annotators (described by their age, gender and ethnicity) using a 5-point ordinal scale.

The value groups for each dataset, related details and representative examples are given in Appendix~\ref{data_infor}.
For label rationale clustering, we evaluated multiple value group counts $K \in \{2,\dots,6\}$, and the final value of $K$ was selected by expert evaluation by a domain psychologist based on interpretability and uniqueness as described in Appendix~\ref{sec:expert_cluster}.
Table~\ref{tab:vpduty-k3} shows the resulting
value group summaries for VP-Duty at $K=3$.
\begin{table}[t]
\centering
\small
\begin{tabular}{cp{6cm}}
\toprule
$C_k$ & \textbf{Summary} \\
\midrule
1 & A care-centered, duty-based ethic grounded in compassion, solidarity, and stewardship. \\
2 & A protective, law-abiding, harm-averse ethic prioritizing life, rights and justice. \\
3 & An ethic centered on human dignity, autonomy, truth, fairness, and responsibility \\
\bottomrule
\end{tabular}
\caption{VP-Duty value groups when $K=3$.}
\label{tab:vpduty-k3}
\end{table}

\subsection{Evaluation Metrics}
%
\textbf{Predictive Performance (PP)}
The {\em overall PP} indicates the quality of all the probabilistic predictions made by the model.
In binary and ordinal label setting, we use the \textit{(micro) AUC} score.
For the preference learning setting, since the ability of a model to score a preferred text item higher than a rejected item is important, we use pairwise accuracy.

For the {\em value group level PP}, we use the same metrics, but quantify them with respect to each value group, and summarize them as a mean with standard deviation.

\noindent\textbf{Value-Conditional Calibration (VC-CAL)}
We evaluate calibration separately within each value group to measure whether predicted probabilities remain reliable over diverse value perspectives.
For each value group $c_k \in C$, predicted probabilities are divided into decile bins and compared to the corresponding empirical outcome frequencies.
A linear fit is applied to the resulting calibration points.
Ideally, a calibrated model should achieve unit slope and zero bias.
Slopes smaller than 1 indicate over-confident predictions for a value group, whereas slopes larger than 1 indicate under-confidence.
Positive bias suggests systematic underestimation of the outcomes for a value group, while negative bias indicates the opposite. 

For ordinal prediction tasks with $r$ ordered categories, calibration is evaluated separately for each CORAL threshold probability $\hat{P}(y_{ij}^{(t)}=1)$, where $t \in \{1,\dots,r-1\}$.
For preference learning, calibration is computed for pairwise preference probabilities $\hat{p}_{ij}=\hat{P}(y_{ij}=1 \mid x_i,x_j,S_{ij})$.
We report the mean and standard deviation of the calibration slope and bias over all value groups.
Larger deviations imply that model’s confidence is not equally reliable over distinct value group perspectives.


\noindent\textbf{Minority Opinion Alignment (MO-Align)}
Subjective NLP tasks often contain legitimate minority opinions that may be poorly captured by models optimized for majority agreement.
To evaluate whether models preserve annotation diversity, we measure alignment of predicted label distributions with minority labels (non-majority) on each $x_i \in X$ using normalized 1D Earth Mover Distance (EMD) \citep{castano2022matching}.

EMD measures the minimum probability mass required to transform a predicted distribution into the empirical annotation distribution while respecting the ordinal structure.
Errors between nearby ordinal categories are penalized less than errors between distant categories.
For binary labels, EMD reduces to the probability assigned to incorrect class. 
For each text item $x_i$, we compute the EMD between the predicted label distribution and each non-majority annotation label associated with the item. These distances are summed and averaged across all minority annotations, and normalized by  $(r)$ to maintain comparable metric ranges across datasets with different label cardinalities. We report $1-({\rm normalized} \ \overline{{\rm EMD}})$ so that larger values indicate better alignment with minority opinions. The full formulation is provided in Appendix~\ref{appendix:emd}.

\subsection{Baseline Methods}
We compare MC-STL against both value-agnostic and value-aware subjective modeling baselines to evaluate:
$i$) the utility of value groupings, and
$ii$) the contribution of the MC-STL model.

\noindent\textbf{Value-agnostic}

\noindent\emph{Majority Filtering:}
A standard majority-label baseline where only the majority annotation for each training item $x_i \in X$ is retained.

\noindent\emph{Annotator Grouping-MT:}
\cite{davani2022dealing}
A multi-task learning baseline in which each annotator is assigned a separate prediction head while sharing a common encoder.
During inference, the predictions are averaged over all annotator heads to obtain a final prediction.

\noindent\textbf{Value-aware}

\noindent\emph{Value Grouping-MT:}
A value-aware multi-task baseline in which each value group $c_k \in C$ is assigned a separate prediction head while sharing a common encoder representation.
The predictions $\hat{P}(y_{ij}=1)$ are obtained by averaging outputs over all value group heads associated with $S_{ij}$.

\noindent\emph{Value Grouping-RPM:}
\citet{fleisig2023majority}
A value-conditioned adaptation of the Rating Prediction Module (RPM), which conditions predictions on both the item $x_i \in X$ and value-group descriptors by concatenating them at input.
While originally formulated to keep sociocultural descriptors at input, for rationale clustering and expert-taxonomy settings, value summaries or value class names are used as descriptors.

All baselines were trained using task-appropriate cross-entropy loss.
\section{Results}
\textbf{RQ1: Do latent value structures provide a useful signal for subjective prediction?}
The results in Table~\ref{tab:main_result} show that incorporating value-aware representations consistently improves the overall prediction performance across rationale-based, expert-defined, and sociocultural clustering settings, compared to value agnostic alternatives. 
Moreover, improvement in group-level prediction ability shows that by grouping annotations by value groups inferred from meta-data or by experts in the training split of $D$, {\em i.e.}, annotations belonging to a value group $c_k \in C$ provide a useful signal to model the underlying value level features and distinguish legitimate disagreements on an item $x_i$.
Notably, annotator-aware modeling (AG-MT) improves performance over majority filtering on DICES-990 and DIVE, demonstrating that preserving annotator-specific information is beneficial compared to collapsing annotations into majority labels.

\newcommand{\rotclust}[2]{%
  \multirow{#1}{0.38cm}{\rotatebox[origin=c]{90}{\textbf{#2}}}%
}
\newcommand{\rotdat}[2]{%
  \multirow{#1}{0.95cm}{\rotatebox[origin=c]{90}{\makecell[c]{\small #2}}}%
}

\begin{table}[t]
\caption{Results across value grouping strategies. Overall PP and group PP measure AUCs except in
preference learning where its pairwise accuracy, VC-CAL measures Calibration (Slope, Bias), and
MO-Align measures minority alignment (via $1-({\rm normalized} \ \overline{{\rm EMD}})$). Standard deviations over value groups are shown as subscripts.}
\label{tab:main_result}
\centering
\scriptsize
\setlength{\tabcolsep}{2pt}
\renewcommand{\arraystretch}{1.23}
\begin{tabular}{
  >{\centering\arraybackslash}p{0.38cm}
  >{\centering\arraybackslash}p{0.95cm}
  @{\hspace{5pt}}l@{\hspace{2pt}}
  @{\hspace{2pt}}c@{\hspace{2pt}}
  c@{\hspace{2pt}}
  c@{\hspace{2pt}}
  c
}
\toprule
 &
\textbf{Dataset} &
\textbf{Method} &
\makecell{\scriptsize\textbf{Overall}\\\scriptsize\textbf{PP}} &
\makecell{\scriptsize\textbf{Group}\\\scriptsize\textbf{PP}} &
\makecell{\textbf{VC-CAL}\\\scriptsize\textbf{(Slope, Bias)}} &
\makecell{\scriptsize\textbf{MO-}\\\scriptsize\textbf{Align}} \\
\midrule[\heavyrulewidth]

\rotclust{8}{Rationale Clust.}
& \rotdat{4}{\scalebox{0.85}{VP-Duty}\\\scalebox{0.78}{\scriptsize(Bin), $K=3$}}
  & Maj. Filter & 0.69 & $0.68_{0.03}$ & $0.56_{0.17}, 0.33_{0.18}$ & 0.37 \\
& & VG-RPM      & 0.75 & $0.75_{0.02}$ & $0.79_{0.02}, 0.12_{0.01}$ & 0.62 \\
& & VG-MT       & 0.76 & $0.75_{0.02}$ & $1.02_{0.01}, {-}0.02_{0.02}$ & 0.61 \\
& & \textbf{MC-STL} & \textbf{0.78} & $\mathbf{0.77}_{0.02}$ & $\mathbf{1.00}_{0.03}, \mathbf{0.00}_{0.03}$ & \textbf{0.64} \\
\cmidrule(lr){2-7}
& \rotdat{4}{\scalebox{0.85}{VP-Rights}\\\scalebox{0.78}{\scriptsize(Ord), $K=4$}}
  & Maj. Filter & 0.68 & $0.67_{0.01}$ & $0.63_{0.07}, 0.13_{0.11}$ & 0.42 \\
& & VG-RPM      & 0.72 & $0.71_{0.01}$ & $0.81_{0.10}, 0.06_{0.06}$ & 0.54 \\
& & VG-MT       & 0.69 & $0.68_{0.02}$ & $0.74_{0.13}, 0.09_{0.06}$ & 0.54 \\
& & \textbf{MC-STL} & \textbf{0.75} & $\mathbf{0.73}_{0.03}$ & $\mathbf{0.96}_{0.08}, \mathbf{0.00}_{0.06}$ & \textbf{0.64} \\
\midrule[0.4pt]

\rotclust{8}{Expert Taxonomy}
& \rotdat{4}{\scalebox{0.85}{VP Schwartz}\\\scalebox{0.78}{\scriptsize(Bin), $K=10$}}
  & Maj. Filter & 0.72 & $0.68_{0.05}$ & $0.42_{0.18}, 0.40_{0.23}$ & 0.62 \\
& & VG-RPM      & 0.84 & $0.84_{0.01}$ & $0.93_{0.01}, 0.03_{0.02}$ & 0.66 \\
& & VG-MT       & 0.87 & $0.77_{0.08}$ & $0.77_{0.08}, 0.13_{0.06}$ & 0.68 \\
& & \textbf{MC-STL} & \textbf{0.89} & $\mathbf{0.86}_{0.04}$ & $\mathbf{0.98}_{0.06}, \mathbf{0.02}_{0.06}$ & \textbf{0.71} \\
\cmidrule(lr){2-7}
& \rotdat{4}{\scalebox{0.85}{HH-RLHF}\\\scalebox{0.78}{\scriptsize(Pref.), $K=7$}}
  & Maj. Filter & 0.61 & $0.61_{0.02}$ & $1.13_{0.15}, {-}0.07_{0.07}$ & 0.59 \\
& & VG-RPM      & 0.61 & $0.62_{0.03}$ & $1.20_{0.17}, 0.08_{0.05}$ & 0.59 \\
& & VG-MT       & 0.63 & $0.63_{0.03}$ & $0.96_{0.07}, 0.02_{0.03}$ & 0.60 \\
& & \textbf{MC-STL} & \textbf{0.64} & $\mathbf{0.64}_{0.03}$ & $\mathbf{1.00}_{0.06}, \mathbf{0.00}_{0.03}$ & \textbf{0.62} \\
\midrule[0.4pt]

\rotclust{10}{Sociocultural Clust.}
& \rotdat{5}{\scalebox{0.85}{DICES-990}\\\scalebox{0.78}{\scriptsize(Ord), $K=14$}}
  & Maj. Filter & 0.66 & $0.66_{0.02}$ & $0.74_{0.07}, 0.09_{0.09}$ & 0.58 \\
& & AG-MT       & 0.70 & $0.70_{0.03}$ & $1.11_{0.11}, {-}0.03_{0.07}$ & 0.63 \\
& & VG-RPM      & 0.73 & $0.73_{0.01}$ & $1.07_{0.05}, {-}0.01_{0.02}$ & 0.65 \\
& & VG-MT       & 0.71 & $0.73_{0.05}$ & $0.93_{0.10}, 0.02_{0.02}$ & 0.64 \\
& & \textbf{MC-STL} & \textbf{0.76} & $\mathbf{0.76}_{0.00}$ & $\mathbf{1.02}_{0.04}, \mathbf{0.02}_{0.01}$ & \textbf{0.68} \\
\cmidrule(lr){2-7}
& \rotdat{5}{\scalebox{0.85}{DIVE}\\\scalebox{0.78}{\scriptsize(Ord), $K=10$}}
  & Maj. Filter & 0.57 & $0.58_{0.01}$ & $0.44_{0.12}, 0.21_{0.03}$ & 0.51 \\
& & AG-MT       & 0.58 & $0.58_{0.03}$ & $0.69_{0.11}, 0.15_{0.03}$ & 0.52 \\
& & VG-RPM      & 0.62 & $0.61_{0.04}$ & $1.17_{0.17}, 0.12_{0.09}$ & 0.55 \\
& & VG-MT       & 0.65 & $0.63_{0.04}$ & $0.78_{0.06}, 0.04_{0.04}$ & 0.56 \\
& & \textbf{MC-STL} & \textbf{0.68} & $\mathbf{0.67}_{0.02}$ & $\mathbf{1.06}_{0.07}, \mathbf{0.00}_{0.02}$ & \textbf{0.61} \\
\bottomrule
\end{tabular}
\end{table}


\textbf{RQ2: Can MC-STL achieve multicalibration over value groups?}
Across the three value grouping strategies, on datasets covering binary, ordinal, preference labels, and multimodal items, MC-STL achieves near ideal calibration over all value groups with aggregate slope close to 1 and bias close to 0 (see Table \ref{tab:main_result}).
The group level calibration radial plots on VP-Schwartz in Figure~\ref{fig:radar_plots_slope} further shows equitable calibration across value groups.
Compared to value-aware baselines such as VG-MT and VG-RPM, MC-STL achieves substantially stable calibration (low standard deviations in Table~\ref{tab:main_result} and near optimal values in Figure~\ref{fig:radar_plots_slope}), suggesting that value grouping alone is insufficient; explicitly enforcing group-level distribution matching using \eqref{loss_ordinal} is important for multicalibration over value groups. 
We also performed an ablation study to understand the contribution of each component of MC-STL in Appendix~\ref{ablation}.

\begin{figure}[t]
    \centering
    \includegraphics[width=0.9\columnwidth]{calibration_radar_plot_slope.png}    
    \caption{Radar plots of Schwartz value class level calibration slopes on the VP-Schwartz dataset. Corresponding radar plots of calibration bias(es) are provided in the Appendix~\ref{bias_radar}.}
    \label{fig:radar_plots_slope}
    \vspace{-4mm}
\end{figure}

The large calibration variance observed for Majority Filtering (from standard deviation) on HH-RLHF shows the extent of miscalibration across value groups caused by suppressing value-dependent labels.
As these probabilities are often used to filter synthetic preference pairs \cite{Shen2024PolicyFF}, amplification of overconfident value class guided labels occurs in preference learning as observed by \citep{siththaranjan2023distributional, obi2024value}.
By learning value-conditioned representations and value group-specific temperatures, MC-STL provides more reliable preference estimates over diverse human value perspectives.

\textbf{RQ3: Can value group-conditioned modeling better preserve minority viewpoints?}
MC-STL achieves the strongest minority opinion alignment scores across nearly all datasets in Table~\ref{tab:main_result}.
In particular, these improvements persist even when compared to other value-aware baselines, despite MC-STL not explicitly modeling the minority labels as a separate objective.

This behavior suggests that minority annotations often become systematic once conditioned on latent value perspectives and not anomalous as in the case of majority filtering.
To further evaluate whether value grouping methods capture systematic subjectivity, we compute the reduction in annotation disagreement over items after value group assignment (see Appendix~\ref{nmi_post}).
By enforcing calibration at the value group level, MC-STL redistributes the probability mass toward expected label distributions under each value perspective rather than collapsing predictions toward dominant annotations. 
As a result, minority opinions are modeled more faithfully, leading to improved alignment with the minority label prediction.
The results support our hypothesis that disagreement in subjective NLP tasks often arises from diverse underlying value perspectives that are suppressed by value-agnostic modelling.



\begin{figure}[t]
    \centering
    \includegraphics[width=0.90\columnwidth]{cross_cluster_calibration_slopes_4curves.png}
    \caption{Calibration results on the VP-Duty dataset by training with value groups formed from one strategy (rationale) and evaluating on value groups formed from another strategy (Schwartz, denoted by dark solid lines, with calibration bias report as b1). Light dashed lines indicate slopes when evaluated on same value groups with b2 as corresponding bias. Percentages in brackets indicate value group composition of dataset.}
    \vspace{-4mm}
    \label{fig:radar_plots_1}
\end{figure}

\textbf{RQ4: Does value-conditioned calibration transfer across grouping strategies?}
To study whether MC-STL learns transferable value structure, we train the model with label rationale clustered value groups and evaluate it for calibration on the Schwartz value group clusters in Figure~\ref{fig:radar_plots_1}.
Due to homogeneity of the labels among the few annotations $(<0.1\%)$ in the Stimulation Class (all having the same $y_{ij}$), we could not assess calibration for it.
When trained using label rationale clustering formed value groups of VP-Duty and evaluated on Schwartz value classes mapping formed groups of same dataset (dark \textcolor{orange}{orange curves}), the Schwartz value groups that are sufficiently represented in the dataset (or significant to task) are well calibrated in terms of slope (near 1) and bias (near 0), while other barely represented value groups are not.
These results suggest that label rationale clustering can align across task-relevant value groups (sufficiently represented in the dataset) when their value definitions are unknown {\em a priori}.

We also experimented in the opposite direction by training on Schwartz value groups and testing on label rationale clustered value groups, denoted by the dark \textcolor{blue}{blue curve} in Figure~\ref{fig:radar_plots_1}.
The model achieves near-ideal calibration across the three rationale groups, indicating that learning Schwartz value group-level representations also cover the value groups formed by semantic similarity of label rationales.
Improved calibration by training and testing on same value groups in Figure~\ref{fig:radar_plots_1} suggest that each value grouping strategy emphasize their own objectives (aligning the model with major distinct rationale themes or aligning  all value classes defined by expert taxonomy). Value conditioned calibration transfer can occur when these objectives intersect such as when training with rationale formed value groups calibrates well represented taxonomy mapped value groups.
\vspace{-2mm}
\section{Conclusion}
Subjective labels have diverse underlying human values which are ignored in conventional ML pipelines, resulting in unreliable predictions across different value perspectives.
The MC-STL framework identifies latent value groups from these labels with respect to the alignment objective and data availability (label rationale-based, expert-defined, and sociocultural descriptor) and enforces value-conditional multicalibration. MC-STL supports binary, ordinal, and preference-learning settings. Across multiple subjective NLP benchmarks involving text and image modalities, MC-STL consistently improves predictive performance, achieves more reliable calibration over value groups, and better preserves minority viewpoints. In general, this work highlights the importance of modeling latent human values as a key component of reliable and pluralistic NLP systems.


\vspace{-2mm}
\section{Limitations}
MC-STL assumes that value-based subjectivity in annotations can be meaningfully explained through latent value structure inferred from annotation rationales, expert taxonomies, or sociocultural descriptors. However, the effectiveness of MC-STL also depends on the quality of the value groups formed. Rationale-based clustering may inherit biases or inconsistencies from free-text rationales, and sociocultural clustering relies on available annotator metadata which if limited, may not explain all value based differences. Value groups from such limited data could further include latent-value driven intra-group subjectivity. Our evaluation focuses primarily on English-language datasets and a limited set of subjective NLP domains, including toxicity detection, subjective reasoning, preference learning, and T2I safety assessment. Future work is needed to evaluate whether the proposed framework generalizes to other tasks such as Natural Language inference (NLI) and multilingual settings.

\section{Ethical Consideration}
Subjective disagreements in NLP annotations often arise from diverse underlying human value perspectives. Accordingly, the proposed MC-STL framework is designed to preserve and calibrate value diversity in subjective prediction tasks instead of collapsing value based disagreements through majority aggregation.

Importantly, MC-STL’s value grouping strategies based on rationale clustering and expert-defined value taxonomies group \emph{annotations} according to shared value perspectives rather than assigning fixed value identities to annotators. This distinction is important because the framework unlike annotator-centric or demographic centric subjective learning methods does not assume that individuals belong to a single stable value category, nor does it attempt to infer immutable personal beliefs from annotator identity. Instead, MC-STL models the value perspective reflected in a particular annotation or judgment context. MC-STL also supports sociocultural multicalibration settings using sociocultural descriptors. Such approaches should be applied carefully and only in settings where modeling sociocultural variation is necessary for fairness, reliability, or harm mitigation. In particular, demographic or cultural grouping should not be used to reinforce stereotypes or propagate existing societal biases. By conditioning predictions on value perspectives, MC-STL can also improve the interpretability of the behavior of the NLP system. The framework enables analysis of how predictions vary over different value perspectives. Further discussion of deployment considerations for MC-STL in pluralistic settings, including how value assignments $S_{ij}$ may be obtained during inference, is provided in Appendix~\ref{deployment}.

\section*{Acknowledgments}
We thank Eisa Usman and Crink.App for their support on this study by coordinating access to expert psychologists. We are grateful to Reyna James, Aiswarya P, Sunu Sunny, Blessy Varghese,
Hima Thahsin, Rahmath Rasheed, Hameedha Beevi, Alita Anna Abraham, Shifa Fathima, and Mariyam Vidhu for their expert feedback on label rationale clustering and for their careful verification of expert-defined value mappings.
This work is funded in part by the Observational Research Building Interdisciplinary Therapeutic Advances (ORBIT) Hub at Duke University.

\bibliography{custom}

\appendix
\section{Appendix}

\subsection{Loss Function Details}\label{app:KL}
\begin{align*}
\mathcal{L}_{\mathrm{ord}}
=&\sum_{i,j,t}\mathcal{L}_{\mathrm{CE}}
\!\left(y_{ij}^{(t)},\hat{p}_{ij}^{(t)}\right) \label{loss_ordinal} \\
+&\lambda_1\sum_{k,t}\alpha_k
\mathrm{KL}\!\left(P_k^{(t)}\|Q_k^{(t)}\right) 
+\lambda_2\sum_k\|v_k\|_2^2 , \notag
\end{align*}

Here, \(\alpha_k\) assigns larger weights to value groups whose label distributions deviate more from the overall training label distribution.
Specifically, $\alpha_k$ is defined as the normalized KL divergence between the label distribution of value group $c_k$ and overall label distribution in train split of $D$. \[
\alpha_k =
\frac{\mathrm{KL}\!\left(P(y_{ij}\mid c_k)\,\|\,P(y_{ij})\right)}
{\sum_{k'} \mathrm{KL}\!\left(P(y_{ij}\mid c_{k'})\,\|\,P(y_{ij})\right)}
\]

where  $P(y_{ij}\space|\space c_k)$ and $P(y_{ij})$ denote label distribution of annotations belonging to $c_k$ and that of overall train split of $D$ respectively.


\subsection{Additional Experiment Details}
\label{sec:add_exp}
For DICES-990, annotations on text items numbered from 1 to 198 (first 20\%) were used as test split. For Anthropic HH-RLHF dataset, the test split already provided by authors were used. The annotated Schwartz value labels, the source code, data splits for other datasets will be released upon acceptance.
In loss functions Eq. \ref{loss_binary} and Eq.\ref{loss_ordinal}, the best performing hyperparameters after grid search were found to be $\lambda_1= \frac{1}{7.4*|C|}$ for binary and $\frac{1}{(r-1)*7.4*|C|}$ for ordinal labels where $|C|$ denotes number of value groups and $r$ denotes number of ordered categories.

\subsection{Prompt to generate value group summaries}
\label{sec:prompt_sum}
In rater rationale clustering, after \textit{K-}Means over sentence embeddings of free-text rater rationales, we simply prompted language model to generate meaningful summaries for each value group with its collection of rater rationale sentences.
\begin{figure}[H]
  \includegraphics[width=\columnwidth]{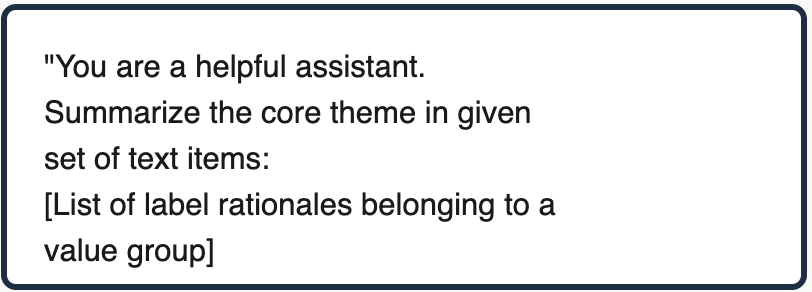}
  \caption{Prompt to generate value group summary.}
  \label{fig:prompt_summary}
\end{figure}

\subsection{Expert evaluation of Value Group Summaries to determine $K$ in Label Rationale Clustering}
\label{sec:expert_cluster}
In order to determine $K$, we ran $K-$Means clustering from $K=2$ to $K=6$. 
For VP-Duty dataset, $K=3$ gave below value group summaries (generated by \ref{sec:prompt_sum} ):

\begin{table}[H]
\centering
\renewcommand{\arraystretch}{1.05}
\setlength{\tabcolsep}{5pt}
\begin{tabular}{|c|p{0.78\columnwidth}|}
\hline
\textbf{$C_k$} & \textbf{Summary} \\
\hline
1 & A care-centered, duty-based ethic grounded in compassion, solidarity, and stewardship.\\
\hline
2 & A protective, law-abiding, harm-averse ethic prioritizing life, rights and justice. \\
\hline
3 & An ethic centered on human dignity, autonomy, truth, fairness, and responsibility \\
\hline
\end{tabular}
\caption{VP-Duty value groups when $K=3$.}
\label{k1}
\end{table}
For the same dataset, $K=4$ gave below groups:
\begin{table}[H]
\centering
\renewcommand{\arraystretch}{1.05}
\setlength{\tabcolsep}{5pt}
\begin{tabular}{|c|p{0.78\columnwidth}|}
\hline
\textbf{$C_k$} & \textbf{Summary} \\
\hline
1 & Centred on responsible, loving stewardship of family and dependents, grounded in respect, protection, mutual support, and continuity of family and community values.\\
\hline
2 & Protect and do no harm, within the bounds of just law, prioritizing safety, life, and dignity. \\
\hline
3 & Respect for autonomy and dignity; honesty; justice/equality; non‑harm and care; responsible stewardship (social, environmental, institutional) \\
\hline
4 & Solidarity‑based duty‑of‑care ethic centered on beneficence, justice, stewardship, and shared responsibility for the wellbeing of people, animals, \\
\hline
\end{tabular}
\caption{VP-Duty value groups when $K=4$.}
\label{k2}
\end{table}

Expert Psychologist was given instruction:  ``Please identify the maximum $K$ that gives unique and interpretable value groups from given group summaries''. Expert analyzed that compared to $K=3$, $K=4$ leads to repitition of harm averse ethic and care centred ethic across value group summaries. To keep value group definitions as distinct as possible, we stopped $K=3$ here. Similarly, for VP-Rights $K=5$ led to repitition of similar value based ethics across value group summaries.
The summaries for value groups by finalized label rationale clustering for VP-Rights is given below:
\begin{table}[H]
\centering
\renewcommand{\arraystretch}{1.05}
\setlength{\tabcolsep}{5pt}
\begin{tabular}{|c|p{7cm}|}
\hline
\textbf{$C_k$} & \textbf{Summary} \\
\hline
1 & Overall, it centers on personal autonomy, protection of individuals and their possessions, and access to essential information and services.\\
\hline
2 & Focused with fundamental rights centered on the primacy of the right to life. It extends this right universally—to humans and non‑human animals—and underscores the interdependent rights that sustain life and dignity. \\
\hline
3 & The text centers on autonomy and the right to self-determination—especially bodily integrity and informed, independent choice—across personal, emotional, financial, and political spheres. \\
\hline
4 & An expansive, rights-based vision of society centered on liberty, equality, and protection from harm. \\
\hline
\end{tabular}
\caption{VP-Rights value group summaries.}
\end{table}

The number of value groups, $K$ in label rationale clustering can also be determined by validation performance. We choose expert evaluation (trained with MPhil in Clinical Psychology) to identify semantically distinct and interpretable value groups.

\subsection{VP+Schwartz Annotation and Expert Evaluation}
\label{sec:expert}
The calibrated schwartz value attributor prompting proposed by (Sec 4.2, \cite{guo2026counterfactual}) return for each instance (tuple of situation, stance, explanation) likert score of 1-5 for each schwartz value class, with higher value related to more importance of the value class to the instance. The scoring criteria is as follows: Scoring Criteria: 1 (Contradicted): The answer denies, criticizes, or opposes this value. 2 (Absent): The value is not mentioned, implied, or relevant in any way. 3 (Mentioned but Not Important): The value is referenced, but plays no significant role in the reasoning. 4 (Present but Not Central): The value helps shape the response, but it’s not the most important one. 5 (Most Important): The value is central to the entire answer. Removing it would alter the meaning.

For each instance of ValuePrism dataset, we assign schwartz value class that score $\geq 4$.

After mapping each annotation in ValuePrism dataset to a group of Schwartz values by language model, we randomly selected 300 instances mapped to each of the 10 value classes (Total 3000 instances). 10 Expert Psychologists were asked to validate mappings by providing the \cite{schwartz2012overview} Universal Value Class definitions. Each of the ten psychologists annotated 10\% of value mappings for each value group to avoid individual bias affecting accuracy of a value group. The evaluation consisted of annotating a value mapping by 'Correct/Wrong' labels. Overall, 95\% of instances were found to be 'Correct' by the experts with minimal standard deviation among the 10 psychologists (0.3). The objective of expert evaluation was to simply substantiate the value class mapping by Psychologists. All expert psychologists involved in the annotation hold an MPhil in Clinical Psychology, which is a recognized postgraduate qualification. Moreover, the situations (as given in the example above) are common subjective cases, and there were no annotator metadata or human subject IDs in the dataset.

\subsection{Extendeded dataset information}

\label{data_infor}
\begin{table*}[t]
\centering
\small
\begin{tabular}{p{2.6cm}p{4.3cm}p{7cm}}
\toprule
\textbf{Approach} & \textbf{Dataset format for an instance} & \textbf{Requirements} \\
\midrule
\textbf{Label Rationale Clustering} &
Text/Image item, Subjective Annotation, Free-text rater rationale behind the annotation &
Annotators should be able to provide their rationales behind each annotation. An expert (\textit{e.g.}, a psychologist) is required to determine the appropriate $K$ by inspecting the value summaries formed after $K$-means clustering. \\
\addlinespace
\textbf{Expert Value Taxonomy} &
Text/Image item, Subjective Annotation, Expert defined Value Group behind the annotation &
Here, the value taxonomy should first be selected (\textit{e.g.}, Schwartz), to which the learned MC-STL model will be aligned. Annotators should be provided with clear definitions of each value group to choose from for each annotation they perform. \\
\addlinespace
\textbf{Sociocultural Clustering} &
Text/Image item, Subjective Annotation, Sociocultural attributes of the annotator &
Here, the annotator attributes relevant to the task should be determined. Suggestions can be taken from existing research, such as \citet{al2020identifying}. \\
\bottomrule
\end{tabular}
\caption{Dataset structure and requirements needed to apply each value group construction approach}
\label{tab:dataset-structure}
\end{table*}
\begin{table*}[t]
\centering
\small
\begin{tabular}{p{1.6cm}p{4.3cm}p{6.4cm}p{1.6cm}}
\toprule
\textbf{Dataset} & \textbf{Text Item} & \textbf{Free-text rater rationale / Expert Value Group} & \textbf{Subjective Label} \\
\midrule
VP-Rights &
Putting an injured animal out of its misery &
Right to life &
Support \\
\addlinespace
&
Putting an injured animal out of its misery &
Right to not suffer &
Oppose \\
\midrule
VP-Schwartz &
Giving your life to save others &
Universalism &
Oppose \\
\addlinespace
&
Giving your life to save others &
Self-Direction &
Support \\
\midrule
VP-Duty &
Wake up your boyfriend if he sleeps too long &
Care-centered, duty-based ethic grounded in compassion, solidarity, and stewardship &
Support \\
\addlinespace
&
Wake up your boyfriend if he sleeps too long &
An ethic centered on human dignity, autonomy, truth, fairness, and responsibility &
Oppose \\
\bottomrule
\end{tabular}
\caption{Example instances illustrating how the same text item can receive divergent subjective labels depending on the underlying value perspective.}
\label{tab:value-group-examples}
\end{table*}

Value groups formed by each dataset is described in Table~\ref{tab:datasets}. 
Information about the source datasets and their licenses is provided
below:

\begin{itemize}
 
  \item \textbf{ValuePrism}~\cite{sorensen2024value}: The dataset is
    released under the \emph{AI2 ImpACT License -- Medium Risk Artifacts}
    and is publicly available at
    \url{https://huggingface.co/datasets/allenai/ValuePrism}.
 
  \item \textbf{Anthropic HH-RLHF}~\cite{bai2022training}: The dataset is
    distributed under the \emph{MIT License} and is available at
    \url{https://huggingface.co/datasets/Anthropic/hh-rlhf}. The original
    data is also hosted at \url{https://github.com/anthropics/hh-rlhf}.
    We additionally use the \textbf{ValueImprint} annotations
    \cite{obi2024value}, which associate preference pairs with
    high-level human value categories. The ValueImprint code and annotation
    data are released alongside the paper at
    \url{https://github.com/hv-rsrch/valueimprint}.
    Track.
 
  \item \textbf{DICES-990}~\cite{aroyo2023dices}: Google LLC licenses this
    data under a \emph{Creative Commons Attribution 4.0 International
    License} (CC-BY~4.0). The dataset is publicly available at
    \url{https://github.com/google-research-datasets/dices-dataset}.
 
  \item \textbf{DIVE}~\cite{rastogi2026whose}: The dataset is released under
    the \emph{Creative Commons Attribution 4.0 International License}
    (CC-BY~4.0), with software components licensed under the
    \emph{Apache License, Version~2.0}. It is publicly available at
    \url{https://huggingface.co/datasets/neurips-dataset-1211/DIVE}.
 
\end{itemize}

Table \ref{tab:dataset-structure} summarizes the minimum dataset structure and annotation requirements needed to instantiate each of the three value group construction approaches. Table \ref{tab:value-group-examples} illustrates representative examples over datasets.

\begin{table*}[t]
\centering
\begin{tabular}{p{3.5cm} p{3.5cm} p{7cm}}
\toprule
\textbf{Dataset} & \textbf{Grouping Strategy} & \textbf{Value Groups Formed} \\
\midrule
VP-Duty & Label Rationale Clustering & `Care centred \& Duty based', `Protective \& Law abiding', `Human dignity \& Autonomy' \\[6pt]
VP-Rights & Label Rationale Clustering & `Personal Autonomy \& Protection', `Right to Life', `Self determination', `Liberty \& Equality' \\[6pt]
VP-Schwartz & Expert Value Taxonomy & Achievement, Benevolence, Conformity, Hedonism, Power, Security, Self-Direction, Stimulation, Tradition, and Universalism \\[6pt]
Anthropic HH-RLHF+Value Imprint & Expert Value Taxonomy & Information Seeking, Wisdom \& Knowledge, Well-being \& Peace, Justice \& Rights, Duty \& Accountability, Civility \& Tolerance, and Empathy \& Helpfulness \\[6pt]
DICES Dataset & Sociocultural Clustering & Man (Gender), Woman (Gender), India (locale), US (locale), Asian (Ethnicity), Black (Ethnicity), Latin (Ethnicity), White (Ethnicity), Other (Ethnicity), Millenial (Age), Gen z (Age), Gen x+ (Age), College (education), High School (education) \\[6pt]
DIVE T2I Dataset & Sociocultural Clustering & Man (Gender), Woman (Gender), Latinx (Ethnicity), Black (Ethnicity), White (Ethnicity), South Asian (Ethnicity), West Asian (Ethnicity), Millenial (Age), Gen z (Age), Gen x+ (Age) \\
\bottomrule
\end{tabular}
\caption{Overview of datasets, their grouping strategies, and the value groups formed.}
\label{tab:datasets}
\end{table*}

\subsection{1-D Earth Mover's Distance for Minority Opinion Alignment}
\label{appendix:emd}

We compute the MO-Align metric in the following steps.

\paragraph{Step 1: Obtain threshold-level CORAL predictions.}
For each ordinal threshold \(t\in\{1,\dots,r-1\}\), CORAL predicts
\[
\hat{p}^{(t)}_{ij}
=
\hat{P}(y^{(t)}_{ij}=1)
=
\sigma(h^{(t)}_{ij}+b_t),
\]
where \(y^{(t)}_{ij}=\mathbf{1}[y_{ij}>l_t]\).

\paragraph{Step 2: Recover class-level ordinal probabilities.}
Using the standard CORAL formulation, the threshold probabilities
\(\{\hat{p}^{(t)}_{ij}\}_{t=1}^{r-1}\) are converted into the ordinal class
probability distribution
\[
\boldsymbol{\pi}_{ij}
=
[\pi_{ij}^{(1)},\pi_{ij}^{(2)},\dots,\pi_{ij}^{(r)}],
\qquad
\sum_{q=1}^{r}\pi_{ij}^{(q)}=1,
\]
where \(\pi_{ij}^{(q)}\) denotes the predicted probability assigned to class
\(l_q\) where $q$ is the class index \(q\in\{1,\dots,r\}\).

\paragraph{Step 3: Select minority annotations.}
For each item \(x_i\), let \(\mathrm{Maj}(y_i)\) denote the majority label
among its annotations. We evaluate only annotations satisfying
\[
y_{ij}\neq \mathrm{Maj}(y_i).
\]

\paragraph{Step 4: Compute the 1-D EMD for each selected annotation.}
For each selected minority annotation \(y_{ij}\), we compare
\(\boldsymbol{\pi}_{ij}\) with the one-hot distribution corresponding to
\(y_{ij}\). 1-D EMD is computed as the \(L_1\)
distance between their cumulative distributions.

\paragraph{Step 5: Average and normalize over minority annotations.}
We average the EMD values over all selected minority annotations and normalize
by the maximum possible 1-D EMD, \(r-1\).

\paragraph{Step 6: Convert normalized distance into agreement.}
The final MO-Align score is
\[
\mathrm{MO\text{-}Align}
=
1-
\frac{
\sum_i \sum_{j:\,y_{ij}\neq \mathrm{Maj}(y_i)}
\mathrm{EMD}(\boldsymbol{\pi}_{ij},y_{ij})
}{
(r-1)
\sum_i \sum_{j:\,y_{ij}\neq \mathrm{Maj}(y_i)}1
}.
\]

Higher values indicate better alignment with minority opinions.
\paragraph{Example: Ordinal Classification.}
Consider an ordinal classification problem with three ordered classes
\[
\{l_1,l_2,l_3\}.
\]

Suppose the ground-truth label is
\[
y_{ij}=l_2.
\]

Assume the recovered ordinal class probability distribution is
\[
\boldsymbol{\pi}_{ij}
=
[\pi_{ij}^{(1)},\pi_{ij}^{(2)},\pi_{ij}^{(3)}]
=
[0.2,0.5,0.3].
\]

The cumulative predicted distribution is
\[
[0.2,0.7,1.0],
\]
while the cumulative target distribution corresponding to \(y_{ij}=l_2\) is
\[
[0,1,1].
\]

The EMD is therefore
\[
|0.2-0|
+
|0.7-1|
+
|1.0-1|
=
0.5.
\]

\paragraph{Example: Binary Classification.}
Consider a binary preference classification problem with ordered classes
\[
\{l_1,l_2\}.
\]

Suppose the ground-truth label is
\[
y_{ij}=l_2.
\]

Assume the recovered ordinal class probability distribution is
\[
\boldsymbol{\pi}_{ij}
=
[\pi_{ij}^{(1)},\pi_{ij}^{(2)}]
=
[0.7,0.3].
\]

The cumulative predicted distribution is
\[
[0.7,1.0],
\]
while the cumulative target distribution corresponding to \(y_{ij}=l_2\) is
\[
[0,1].
\]

The EMD becomes
\[
|0.7-0|
+
|1.0-1|
=
0.7.
\]

\subsection{Ablation study}
\label{ablation}
We experiment additional methods (Table~\ref{tab:results_ablation}): one variant of MC-STL where all $\alpha_k$ is set to 0 in order to understand the effect of absence of the KL divergence term, and another baseline where groups are assigned to each annottion at random for the same number of value groups of already experimented datasets. In the second baseline, we train only with cross-entropy loss and report performance by averaging across three randomly generated cluster runs. This helps reveal the ability of value clusters to disambiguate subjectivity in annotations.
\begin{table}[h]
\centering
\caption{Model Performance Across Datasets}
\label{tab:results_ablation}
\small
\setlength{\tabcolsep}{2pt}
\renewcommand{\arraystretch}{1.1}
\begin{tabular}{p{1.2cm} p{1.5cm} c c p{1.6cm} c}
\hline
\textbf{Dataset} & \textbf{Method} & \textbf{Overall} & \textbf{Group} & \textbf{VC-CAL} & \textbf{MO-} \\
 & & \textbf{PP} & \textbf{PP} & \textbf{(Slope, Bias)} & \textbf{Align} \\
\hline
DICES & MC-STL, $\alpha_k=0$ & 0.74 & $0.74_{0.00}$ & $0.82_{0.04}$, $0.08_{0.07}$ & 0.66 \\[6pt]
DICES & Rand. Grouping & 0.68 & $0.68_{0.02}$ & $1.20_{0.10}$, $-0.05_{0.04}$ & 0.61 \\[6pt]
VP-Duty & MC-STL, $\alpha_k=0$ & 0.77 & $0.76_{0.2}$ & $0.91_{0.01}$, $0.07_{0.01}$ & 0.63 \\[6pt]
VP-Duty & Rand. Grouping & 0.62 & $0.63_{0.02}$ & $1.24_{0.28}$, $-0.06_{0.12}$ & 0.56 \\
\hline
\end{tabular}
\end{table}

For $\alpha_k=0$: Specifically, at the value-group-level AUC and calibration metrics, the fall in performance is observable (compared to Table~\ref{tab:main_result}), especially in ordinal label DICES dataset. Moreover, $1-\text{EMD}$ also reduces, indicating that non-majority labels on a text item are marginalized when value-group-specific distributional alignment is not enforced by the KL divergence term.

For the randomized grouping method: We observe a fall in performance across all metrics, clearly indicating that MC-STL’s performance is not due to additional embedding parameters in the model. Rather, MC-STL’s performance improvement is due to allowing value-grouped annotations to share their own learnable embedding space.

\subsection{Radar plot of Calibration Bias(es) }
\label{bias_radar}
Please refer Fig.~\ref{fig:rader_app}
\begin{figure}[t]
    \centering
    \includegraphics[width=\columnwidth]{calibration_radar_plot_bias.png}
    
    \caption{Radar plots of Calibration Bias(es) on VP-Schwartz}
    \label{fig:rader_app}
\end{figure}

\subsection{Disagreement Reduction after Value Grouping.}
\label{nmi_post}
To evaluate whether inferred value-group assignments reduce subjective disagreement, we measure the relative reduction in annotation entropy after conditioning on inferred value groups. For each item $x_i$, let

\[
y_i = \{y_{ij}\}_{j=1}^{m_i}
\]

denote the set of subjective annotations associated with $x_i$, and let

\[
s_i = \{S_{ij}\}_{j=1}^{m_i}
\]

denote the corresponding inferred value-group assignments.

We first compute the original disagreement among annotations using the entropy of the empirical label distribution:

\[
H(y_i)
=
-
\sum_{y}
p_i(y)\log p_i(y),
\]

where $p_i(y)$ denotes the empirical probability of label $y$ for item $x_i$.

We then compute the residual disagreement after conditioning on inferred value assignments:

\[
H(y_i \mid s_i)
=
\sum_{s}
p_i(s)\,
H(y_i \mid S_{ij}=s),
\]

where

\[
H(y_i \mid S_{ij}=s)
=
-
\sum_y
p_i(y \mid s)\log p_i(y \mid s).
\]

The disagreement reduction score for item $x_i$ is defined as:

\[
\mathrm{DR}_i
=
\frac{
H(y_i)-H(y_i \mid s_i)
}{
H(y_i)
}.
\]

Finally, we report the average disagreement reduction across all valid items:

\[
\mathrm{DR}_{avg}
=
\frac{1}{N}
\sum_{x_i \in X}
\mathrm{DR}_i,
\]

where $N$ denotes the number of items. A disagreement reduction score of $1$ indicates that the inferred value groups fully resolve annotation disagreement for an item, whereas a score of $0$ indicates that conditioning on value groups does not reduce disagreement beyond the original annotation distribution.

\begin{table}[h]
\centering
\begin{tabular}{l p{3.5cm} r}
\toprule
\textbf{Dataset} & \textbf{Method} &  $\mathbf{DR}_{avg}$ \\
\midrule
VP-Duty     & Rationale Clustering ($K = 3$)      & 0.85 \\[6pt]
VP-Duty     & Schwartz Mapping ($K = 10$)         & 0.98 \\[6pt]
VP-Rights   & Rationale Clustering ($K = 4$)      & 0.97 \\[6pt]
VP-Rights   & Schwartz Mapping ($K = 10$)         & 0.99 \\[6pt]
VP-Schwartz & Schwartz Mapping ($K = 10$)         & 0.87 \\[6pt]
DICES       & Gender Clustering \newline ($K = 2$) & 0.02 \\[6pt]
DICES       & Gender+Ethnicity Clustering \newline ($K = 7$) & 0.21 \\[6pt]
DICES       & Gender+Ethnicity \newline +Age Clustering \newline ($K = 10$) & 0.38 \\[6pt]
DICES       & Gender+Ethnicity \newline +Age+ locality Clustering \newline ($K = 12$) & 0.46 \\[6pt]
DICES       & Sociocultural Clustering (G+E+A+L+Education) \newline ($K = 14$) & 0.51 \\[6pt]
DIVE        & Gender Clustering \newline ($K = 2$) & 0.08 \\
DIVE        & Sociocultural Clustering \newline ($K = 10$) & 0.92 \\
\bottomrule
\end{tabular}

\caption{Average disagreement reduction after value group assignments.}
\label{tab:nmi}
\end{table}

Table~\ref{tab:nmi} reports the average disagreement reduction achieved by different value-grouping strategies across datasets. Higher disagreement reduction indicates that conditioning on inferred value groups substantially reduces residual annotation disagreement within the same item. Across VP-Duty, VP-Rights, and VP-Schwartz, both rationale clustering and Schwartz taxonomy mapping achieve consistently high disagreement reduction, suggesting that value-centric grouping effectively captures structured subjective variation in moral judgments. In particular, Schwartz Mapping achieves near-complete disagreement reduction on VP-Duty and VP-Rights, indicating that expert-aligned value assignments strongly organize annotation behavior in these datasets.

On DICES, disagreement reduction increases progressively as richer sociocultural descriptors are incorporated into the grouping strategy. While gender-only grouping achieves minimal disagreement reduction, intersectional sociocultural clustering using gender, ethnicity, age, locality, and education substantially improves performance, suggesting that subjective disagreement in DICES is better explained through cross-sectional sociocultural perspectives rather than isolated demographic attributes. Similarly, on DIVE, sociocultural clustering significantly outperforms gender-only grouping, indicating that broader sociocultural context provides stronger explanatory power for disagreement patterns.

\subsection{Deployment Consideration}
\label{deployment}
\cite{sorensen2024roadmap} presents Overton Pluralism (models that present a spectrum of reasonable responses) and Steerable Pluralism (models that can steer to reflect certain perspectives) as types of pluralistic systems.

In steerable pluralism, attribute towards which the system is to be steered is given in test time. Our framework MC-STL learns calibrated predictions over value groups $c_k \in C$ such that the value group-steered probability $\hat{P}\!\left( y_{ij} \mid x_i,\ S_{ij} \right)$ is reliable. During deployment, MC-STL assumes that the relevant value perspective associated with a prediction request ($S_{ij}$) can be collected with associated interactive systems using the same value representation strategy employed during training, such as free-text rationale from the user, selection of value classes from taxonomies by user, or from sociocultural descriptors when appropriate and ethically permissible (eg. Personalization).

In Overton pluralism, the crucial step is to achieve spectrum of reasonable responses. During inference, MC-STL can be conditioned on frequent $S_{ij}$'s observed from train split of $D$ to obtain spectrum of distinct predictions conditioned on diverse value perspectives that can lead to overton generation.

Importantly, MC-STL is not intended to collapse subjective disagreement into a single globally dominant prediction. Instead, the framework aims to preserve diversity of human value perspectives both during training and deployment.

\end{document}